\def\year{2022}\relax
\documentclass[letterpaper]{article} % DO NOT CHANGE THIS
\usepackage{aaai22}  % DO NOT CHANGE THIS
\usepackage{times}  % DO NOT CHANGE THIS
\usepackage{helvet}  % DO NOT CHANGE THIS
\usepackage{courier}  % DO NOT CHANGE THIS
\usepackage[hyphens]{url}  % DO NOT CHANGE THIS
\usepackage{graphicx} % DO NOT CHANGE THIS
\usepackage{natbib}  % DO NOT CHANGE THIS AND DO NOT ADD ANY OPTIONS TO IT
\usepackage{caption} % DO NOT CHANGE THIS AND DO NOT ADD ANY OPTIONS TO IT
\DeclareCaptionStyle{ruled}{labelfont=normalfont,labelsep=colon,strut=off} % DO NOT CHANGE THIS
\usepackage{algorithm}
\usepackage{algorithmic}

\usepackage{amsmath}
\usepackage{amssymb}
\usepackage{booktabs}
\usepackage{multirow}
\usepackage{bbding}
\usepackage{diagbox}
\usepackage{appendix}
\usepackage{subfigure}

\usepackage{newfloat}
\usepackage{listings}
\floatstyle{ruled}
\newfloat{listing}{tb}{lst}{}
\floatname{listing}{Listing}
\title{Spikeformer: A Novel Architecture for Training High-Performance 
\\Low-Latency Spiking Neural Network}
\author{
    Yudong Li,
    Yunlin Lei,
    Xu Yang\thanks{Corresponding author}
}
\affiliations{
    School of Computer Science and Technology, Beijing Institute of Technology, Beijing 100081, China\\
    \{loyd, leiyunlin, pyro\_yangxu\}@bit.edu.cn
}

\begin{document}

\maketitle

\begin{abstract}
    Spiking neural networks (SNNs) have 
    made great progress on both performance and efficiency over the last few years, 
    but their unique working pattern makes it hard to train a high-performance low-latency 
    SNN. Thus the development of SNNs still lags behind traditional artificial neural networks (ANNs). 
    To compensate this gap, many extraordinary works have been proposed. Nevertheless, these works 
    are mainly based on the same kind of network structure (i.e. CNN) and 
    their performance is worse than 
    their ANN counterparts, which limits the applications of SNNs. To this end, 
    we propose a novel 
    Transformer-based SNN, termed ``Spikeformer'', which outperforms its 
    ANN counterpart on both static dataset and neuromorphic dataset and may be an alternative 
    architecture to CNN for training high-performance SNNs. 
    First, to deal with the problem of ``data hungry'' 
    and the unstable training period exhibited in the vanilla model, we design the 
    Convolutional Tokenizer (CT) module, which improves the accuracy of the 
    original model on DVS-Gesture by more than 16\%. Besides, in order 
    to better incorporate the attention mechanism inside Transformer and the 
    spatio-temporal information inherent to SNN, we adopt spatio-temporal attention (STA) 
    instead of spatial-wise or temporal-wise attention. With our proposed method, we achieve 
    competitive or state-of-the-art (SOTA) SNN performance on DVS-CIFAR10, 
    DVS-Gesture, and ImageNet datasets with the least simulation time steps (i.e. low latency). 
    Remarkably, our Spikeformer outperforms other SNNs on ImageNet by a large margin (i.e. more than 
    5\%) and even outperforms its ANN counterpart 
    by 3.1\% and 2.2\% on DVS-Gesture and ImageNet respectively, indicating that Spikeformer 
    is a promising architecture for training large-scale SNNs and may be more 
    suitable for SNNs compared to CNN. 
    We believe that this work shall keep the development of SNNs 
    in step with ANNs as much as possible. Code will be publicly available.
    \end{abstract}
    \section{Introduction}
    Inspired by biological neural functionality, spiking neural network (SNN) is known 
    as a promising bionic model with low-power computation and has gained considerable attention 
    in recent years. As the third generation of artificial neural network (ANN), 
    SNN is computationally more powerful than other neural networks with regard to 
    the number of neurons that are needed \cite{maass1997networks}. 
    When embedded on neuromorphic 
    hardware such as TrueNorth \cite{debole2019truenorth} and Loihi 
    \cite{davies2018loihi}, 
    SNN shows great potential to process spatio-temporal information effectively with 
    low energy consumption due to its intrinsic spatio-temporal characteristic and 
    event-driven spike communication mechanism. Nevertheless, owing to the lack of appropriate 
    learning algorithm, its performance is not as good as traditional neural networks.
    
    To tackle this problem, there are two main routes to train a SNN with high 
    performance. The first is the ANN-to-SNN method, which obtains a SNN by 
    simulating a pre-trained ANN model's behavior 
    \cite{hu2018spiking, han2020rmp, sengupta2019going}. 
    With adequate simulation steps, the conversion 
    method can achieve almost lossless accuracy compared to its ANN counterpart.
    However, extremely high latency is often required to achieve satisfying accuracy and 
    limits the practical applications of SNNs. 
    The other approach is the surrogate gredient (SG) method \cite{neftci2019surrogate}. 
    To handle the non-differentiability of the spiking mechanism, the SG method 
    utilizes surrogate gradients to approximate the gradients of the non-differentiable 
    activation function on the backpropagation process. This method can directly train 
    SNNs with low latency, but it cannot achieve high performance comparable to 
    leading ANNs. Overall, SNNs obtained by aforementioned methods suffer from 
    either low performance or high latency which hinders the development of training 
    large-scale SNN models.
    
    To deal with it, 
    there are several methods and models proposed to train large-scale SNNs, such 
    as tdBN \cite{zheng2021going} and SEW-ResNet \cite{fang2021deep}. 
    These methods and models are mainly based on convolutional 
    neural networks (CNNs), such as ResNet \cite{he2016deep} and 
    VGG \cite{Simonyan2015VeryDC}, 
    which dominate the field of computer vision for years. 
    However, in the field of computer vision, CNNs are gradually taken place by Vision
    Transformers (ViT) because of their excellent capabilities at capturing long-range
    dependencies. With the emergence of Transformer \cite{vaswani2017attention} and 
    ViT \cite{dosovitskiy2020image}, 
    they have been the dominant approaches in sequence modeling problems such as language 
    modeling and video understanding. Actually, 
    the tasks of SNNs (e.g. DVS-Gesture) can be viewed as sequence modeling problems as well. 
    Hence, it's natural to integrate SNN with inherent spatio-temporal information into 
    ViT with spatio-temporal attention (e.g. TimeSformer). However, due to the 
    ``data hungry'' property and the substandard optimizability of ViT, 
    the vanilla architecture suffers from poor performance (82.29\% on DVS-Gesture) and the
    training accuracy even does not converge when integrated with spiking neurons. 
    
    To this end, we design the Convolutional Tokenizer (CT) module, which is a SNN-oriented 
    convolutional block. With this module, we improve the top-1 accuracy of the 
    original model on DVS-Gesture from 82.29\% to 95.83\%. And its SNN counterpart 
    manages to converge and even obtains better result (i.e. 98.96\% on DVS-Gesture). To our 
    best knowledge, this is the first work to introduce spatio-temporal attention (STA) into 
    SNNs and directly train a Transformer-based SNN with high performance and low latency. 
    Although there are a few works utilizing attention mechanism in SNNs 
    to improve performance, they are mainly studying either 
    spatial-wise attention \cite{cannici2019attention, xie2016efficient, kundu2021spike} 
    or temporal-wise attention \cite{yao2021temporal} separately. On the contrary, we 
    integrate spatial and temporal attention into SNNs at the same time and achieve 
    state-of-the-art SNN performance on DVS-Gesture, DVS-CIFAR10 and ImageNet datasets with the 
    least simulation time steps (i.e. low latency). Prior to this work, \cite{mueller2021spiking} 
    obtain a Transformer-based SNN by conversion method. But their method requires 
    relatively long time steps and they only experiment on simple tasks, while ours demands 
    much fewer simulation time steps and achieves remarkable result on the challenging ImageNet dataset. 
    On ImageNet, our Spikeformer obtains 7\%/5\% improvement compared 
    to the SOTA SNN models trained by SG/conversion method and even outperforms its ANN counterpart 
    by 2.2\% for the first time.
    
    We summarize our contributions as follows:
    \begin{itemize}
        \item To deal with the ``data hungry'' property and substandard optimizability of ViT, we design the CT module and gain essential improvement of accuracy (over 16\%) on DVS-Gesture.
        \item We integrate STA into SNN to better utilize the spatio-temporal information inside the spiking neurons and achieve high performance with low latency.
        \item We propose a directly trained Transformer-based SNN, which may be an alternative scheme to CNN for training large-scale SNNs.
        \item With extensive experiments, we demonstrate the robustness of Spikeformer in terms of depth and time steps and achieve competitive or state-of-the-art (SOTA) SNN performance with low latency on DVS-CIFAR10, DVS-Gesture, and ImageNet datasets.
    \end{itemize}
    
    \section{Related Work}
    \subsubsection{Large-scale SNN model}
    In the past few years, as SNNs have attracted increasing attention, lots of extraordinary
    works on training large-scale SNN models have sprung up. \cite{hu2018spiking} 
    are the first to 
    build a SNN deeper than 100 layers by conversion method. But it requires hundreds or 
    thousands of time steps, which results in high latency. It is not until 
    \cite{zheng2021going} 
    propose the tdBN method that we can extend directly-trained SNNs from fewer than 10 
    layers to 50 layers. They effectively alleviate gradient vanishing or explosion and 
    balance the threshold and input on each neuron to get appropriate firing rates with 
    a modified batch normalization method. \cite{fang2021deep} further analyze the gradient and 
    identity map issues in SNNs, alter the connection of residual block, and propose 
    SEW ResNet, which extends the depth to 152 layers.  
    These large-scale SNNs are mainly based on CNNs, while Transformer-based architectures have gained 
    great success in various domains (e.g. NLP, CV).
    
    \subsubsection{Transformer}
    Since the proposal of Transformer \cite{vaswani2017attention}, it has been a dominant 
    approach in NLP due to its extraordinary capabilities at capturing long-range 
    dependencies among words. \cite{dosovitskiy2020image} are the first to apply it in computer vision 
    and achieve remarkable results, which sets off a wave of research on ViT. 
    \cite{dosovitskiy2020image} note that ViT lacks inductive biases inherent to CNN, and 
    therefore it requires more data to gain superior results. This issue has subsequently 
    led to a series of studies to follow 
    \cite{liu2021efficient, mehta2021mobilevit, lee2021vision, hassani2021escaping, 
    cao2022training}. 
    \cite{hassani2021escaping} introduce 
    inductive biases into ViT with a convolutional head 
    and successfully alleviate ``data hungry''. 
    Moreover, \cite{xiao2021early} propose that ViT exhibits substandard optimizability 
    due to the patchify stem of ViT.
    They empirically demonstrate that early convolution can increase optimization 
    stability of ViT and also improve peak performance. Despite the rapid development 
    of Transformers in the field of ANNs, they are rarely used in SNNs due 
    to the ``data hungry'' property and the substandard optimizability. \cite{mueller2021spiking} 
    convert a pre-trained Transformer to SNN, but this method requires 
    relatively long time steps and has limited usages. Based on these observations, 
    we propose the CT module and successfully train a high-performance Transformer-based SNN 
    with low latency.
    \subsubsection{Attention mechanism in SNN}
    The attention mechanism enables the model to pay more attention to the most 
    informative components of input. It has been applied in SNN as spatial-wise 
    attention \cite{cannici2019attention, xie2016efficient, kundu2021spike} 
    or temporal-wise attention 
    \cite{yao2021temporal}, but these works either capture dependencies inside frames or 
    utilize the statistical characteristics of the frames input at
    different time steps, which may lose information in spatial or temporal domain. 
    Thus we integrate both spatial and temporal attention into SNN in a divided way and achieve 
    state-of-the-art results.

    \section{Method}

    \begin{figure}[t]
        \centering
        \includegraphics[width=0.9\columnwidth]{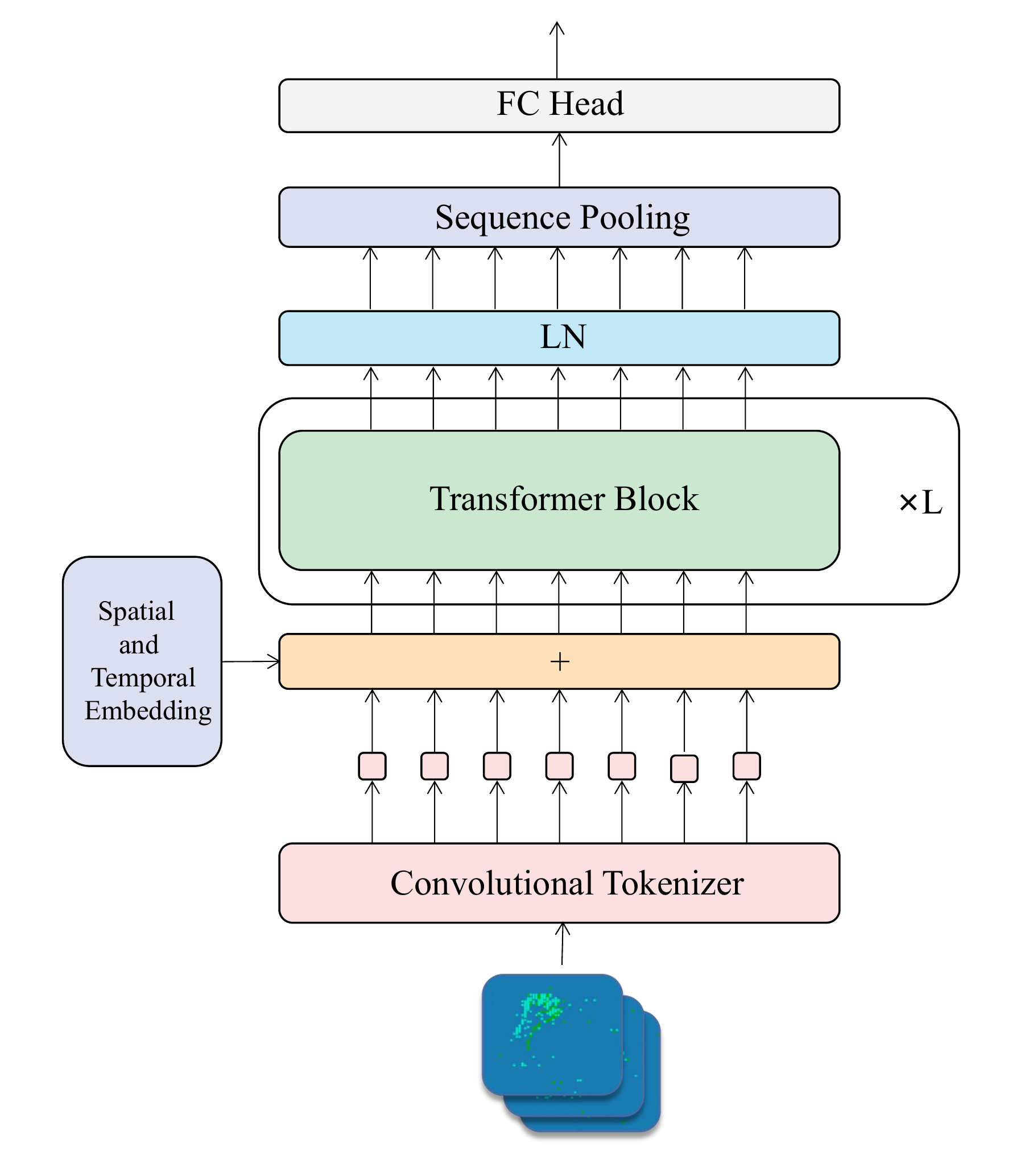}
        \caption{An overview of Spikeformer.}
        \label{fig1}
    \end{figure}

    \subsection{Spiking Neuron Model}
    In general, spiking neurons can be clssified into two classes by their data 
    transmission form. One is the spike-based neuron, such as LIF \cite{burkitt2006review} 
    and PLIF \cite{fang2021incorporating}, 
    which uses spike streams for communication. When embedded on 
    neuromorphic hardware, this kind of neurons will skip computation if they haven't 
    received any input spike, i.e. event-driven computation. Thus they can process 
    spatio-temporal information in an energy-saving way. The other is the analog-based 
    neuron, such as LIAF \cite{wu2021liaf}, which uses the dynamic characteristics of spiking neurons 
    but transmits analog values. This kind of neurons possesses strong representational 
    capabilities due to their inherent spatio-temporal feature, but they cannot skip 
    computation, which makes them cost more energy than spike-based neurons. Without 
    loss of generality, we experiment with both types of neurons.
    Formally, We use the following equations 
    to describe the dynamics of spiking neurons, 
    % \begin{equation}
    %     H[t] = f(V[t-1], X[t]),
    % \end{equation}
    \begin{equation}
        H[t] = V[t - 1] + \frac{1}{\tau}(X[t] - (V[t - 1] - V_{reset}))
        \label{eq1}
    \end{equation}
    \begin{equation}
        S[t] = \Theta(H[t] - V_{th}),
        \label{eq2}
    \end{equation}
    \begin{equation}
        V[t] = H[t](1-S[t]) + V_{reset}S[t],
    \end{equation}
    % \begin{equation}
    %     X[t]=\left\{
    %     \begin{array}{rcl}
    %     S[t] & & {for\ LIF}\\
    %     ReLU(H[t]) & & {for\ LIAF}
    %     \end{array} \right.
    % \end{equation}
    where $H[t]$ and $V[t]$ denote the membrane potential after integrating 
    input and after the trigger of a spike at time step $t$, respectively. $X[t]$ is the input at 
    time-step $t$, and $S[t]$ is the output spike at time-step $t$. 
    The input 
    propagates to the next layer is $S[t]$ for LIF model and $ReLU(H[t])$ for LIAF model. 
    $\Theta(x)$ is the Heaviside 
    step function, which is defined by $\Theta(x) = 1$ for $x > 0$, otherwise 
    $\Theta(x) = 0$. $V_{th}$ and $V_{reset}$ denote firing threshold and reset 
    potential, respectively. $\tau$ represents the membrane time constant. 
    When $\tau$ is learnable, we have PLIF model \cite{fang2021incorporating} and PLIAF model. In this paper, the surrogate 
    method is used in the backpropagation process, and the surrogate function is the same 
    as \cite{fang2021deep}. The details are presented in \textbf{Supplementary Material A}.

    \subsection{Model Architecture}
    An overview of Spikeformer is depicted in Fig.~\ref{fig1}. Compared to the standard ViT, we use a 
    Convolutional Tokenizer (CT) module to process a series of 2D frames and 
    then reshape them into a sequence of flattened token embeddings instead of 
    the original patchify stem. Furthermore, 
    to better utilize the spatio-temporal information and achieve a trade-off between accuracy 
    and computational complexity, we adopt divided Space-Time Attention \cite{bertasius2021space} 
    in Transformer block. Finally, as it is non-trivial to deal with class token in 
    divided Space-Time Attention and the class token may lose information when discarding 
    embedings \cite{beyer2022better}, we pool the sequential based information from the last 
    transformer block using a sequential pooling method \cite{hassani2021escaping}, 
    which may be a better choice than global average-pooling as it can learn to assign more 
    weights to more informative tokens. We would elaborate each component 
    in the following subsections.

    \begin{figure}[h]
        \centering
        \includegraphics[width=0.6\columnwidth]{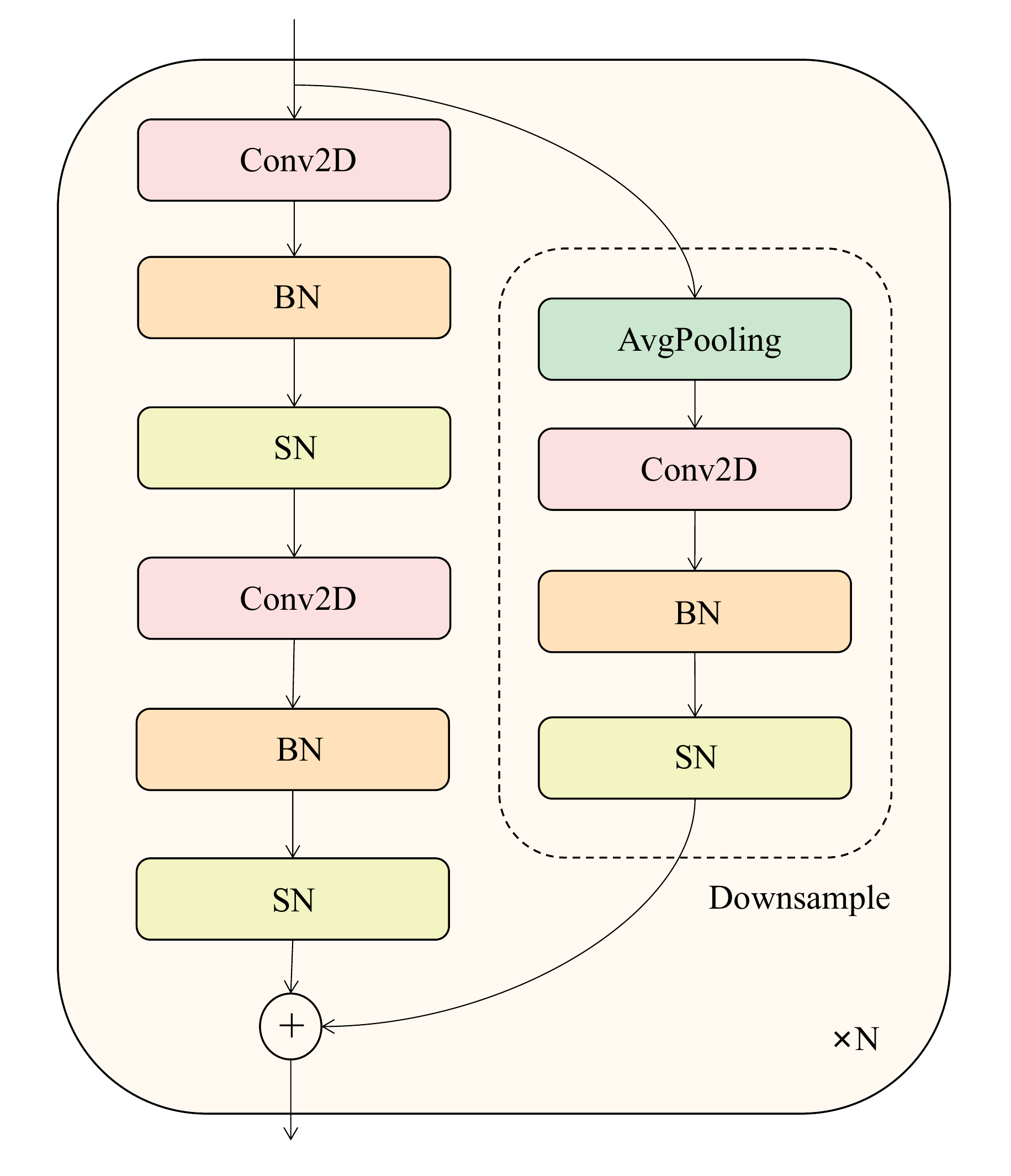}
        \caption{Illustration of Convolutional Tokenizer. Note that the part enclosed 
        by the dotted line only appears when downsampling.}
        \label{fig2}
    \end{figure}
    
    \subsubsection{Convolutional Tokenizer}
    In order to alleviate ``data hungry'' and help stabilize the training period, 
    we propose the CT module to introduce inductive biases into 
    Spikeformer. The architecture of CT module is illustrated in Fig.~\ref{fig2}.
    Given a series of input frames $x\in\mathbb{R}^{T\times H\times W\times C}$:
    \begin{equation}
        Conv(x) = SN(BN(Conv2d(x))),
    \end{equation}
    \begin{equation}
        CT_d(x) = Conv(Conv(x)) + Conv(AvgPool(x)),
    \end{equation}
    \begin{equation}
        CT(x) = Conv(Conv(x)) + x,
    \end{equation}
    where Conv2d and BN denote convolutional layers and batch normalization layers, 
    which will process each time step separately, and SN denotes spiking 
    neurons that will integrate input and output spike trains. $CT_d(\cdot)$ and $CT(\cdot)$ 
    depict downsample module and normal module respectively. Multiple downsample modules and 
    normal modules are stacked together to encode input frames and output spike trains to the following 
    blocks. The design of the CT module mainly follows the architecture 
    of SEW block \cite{fang2021deep} 
    as it is proven to implement identity mapping and is beneficial for training 
    deep spiking neural networks. But in the downsample module, we use a $2\times2$ average 
    pooling layer with a stride of 2 and a stride-1 Conv$1\times1$ instead of a stride-2 Conv$1\times1$ 
    in the shortcut connection in order not to discard information \cite{he2019bag}. And this 
    minimal modification will not influence the functionality of identity mapping, the theoretical 
    analysis and further description are presented in \textbf{Supplementary Material A}.

    \begin{figure}[h]
        \centering
        \includegraphics[width=0.8\columnwidth]{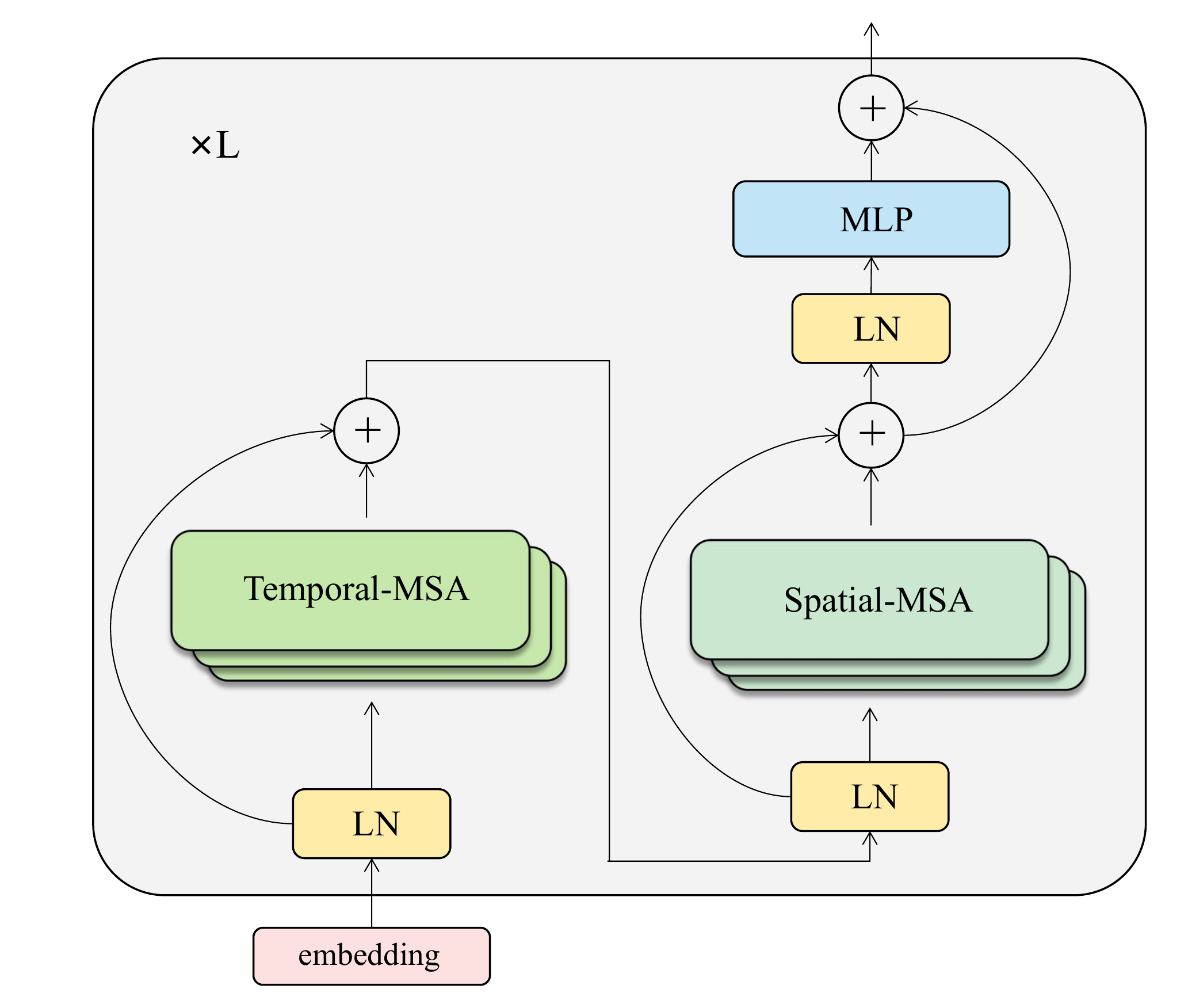}
        \caption{Illustration of Transformer Block.}
        \label{fig3}
    \end{figure}
    
    \subsubsection{Transformer Block}
    Our Spikeformer contains $L$ encoding blocks and the architecture of transformer block 
    is shown in Fig.~\ref{fig3}. The output of the last CT module 
    $\hat{x} \in \mathbb{R}^{T\times H'\times W'\times D}$ is flattened into 
    vectors $\hat{a}_{(p, t)}\in \mathbb{R}^{D}$, with $p=1,...,N$ denoting spatial locations and
    $t=1,...,T$ denoting temporal locations. $N=H'\times W'$ and $D$ depicts the dimensionality 
    of tokens. To encode the spatio-temporal position of each token, we add a learnable 
    positional embedding $e^{pos}_{(p,t)}\in \mathbb{R}^D$ to each token:
    \begin{equation}
        z^{(0)}_{(p, t)} = \hat{a}_{(p, t)} + e^{pos}_{(p, t)}
    \end{equation}
    
    Each encoding block $l$ will compute a query/key/value vector:
    \begin{equation}
        q^{(l,h)}_{(p,t)} = LN(z^{(l)}_{(p,t)})W^{(l, h)}_Q \in \mathbb{R}^{D_h}
    \end{equation}
    \begin{equation}
        k^{(l,h)}_{(p,t)} = LN(z^{(l)}_{(p,t)})W^{(l, h)}_K \in \mathbb{R}^{D_h}
    \end{equation}
    \begin{equation}
        v^{(l,h)}_{(p,t)} = LN(z^{(l)}_{(p,t)})W^{(l, h)}_V \in \mathbb{R}^{D_h}
    \end{equation}
    where $LN(\cdot)$ denotes LayerNorm \cite{ba2016layer}, $h=1,...,H$ is the index over multiple 
    attention heads and $D_h=D/H$ is the dimensionality of each query/key/value vector. 
    In Spikeformer, the attention mechanism is divided 
    into spatial attention and temporal attention:
    \begin{equation}
        h^{(l,h)space}_{(p,t)} = \sum_{p'=1}^{N}{SM(\frac{q^{(l,h)^T}_{(p,t)}}{\sqrt{D_h}}
        \begin{bmatrix}\{k^{(l,h)}_{(p',t)}\}_{p'=1,...,N}\end{bmatrix})v^{(l,h)}_{(p',t)}},
    \end{equation}
    \begin{equation}
        h^{(l,h)time}_{(p,t)} = \sum_{t'=1}^{T}{SM(\frac{q^{(l,h)^T}_{(p,t)}}{\sqrt{D_h}}
        \begin{bmatrix}\{k^{(l,h)}_{(p,t')}\}_{t'=1,...,T}\end{bmatrix})v^{(l,h)}_{(p,t')}},
    \end{equation}
    \begin{equation}
        s^{(l)}_{(p,t)}=
        Concat(h^{(l,1)}_{(p,t)},...,h^{(l,H)}_{(p,t)})\cdot W_O,
    \end{equation}
    where $SM(\cdot)$ denotes the softmax activation function, $W_O \in \mathbb{R}^{H\cdot D_h\times D}$ 
    is the projection matrix, and $s$ is the output of 
    temporal attention (TMSA) or spatial attention (SMSA). Thus a transformer block in 
    Spikeformer can be formulated as:
    \begin{equation}
        z'_l = TMSA(LN(z_{l-1})) + z_{l-1}
    \end{equation}
    \begin{equation}
    z''_l = SMSA(LN(z'_l)) + z'_l
    \end{equation}
    \begin{equation}
        z_l = FC(SN(FC(LN(z''_l)))) + z''_l
    \end{equation}
    with $l=1,...,L$ depicting an index over blocks.
    
    \begin{figure}[h]
        \centering
        \includegraphics[width=1.0\columnwidth]{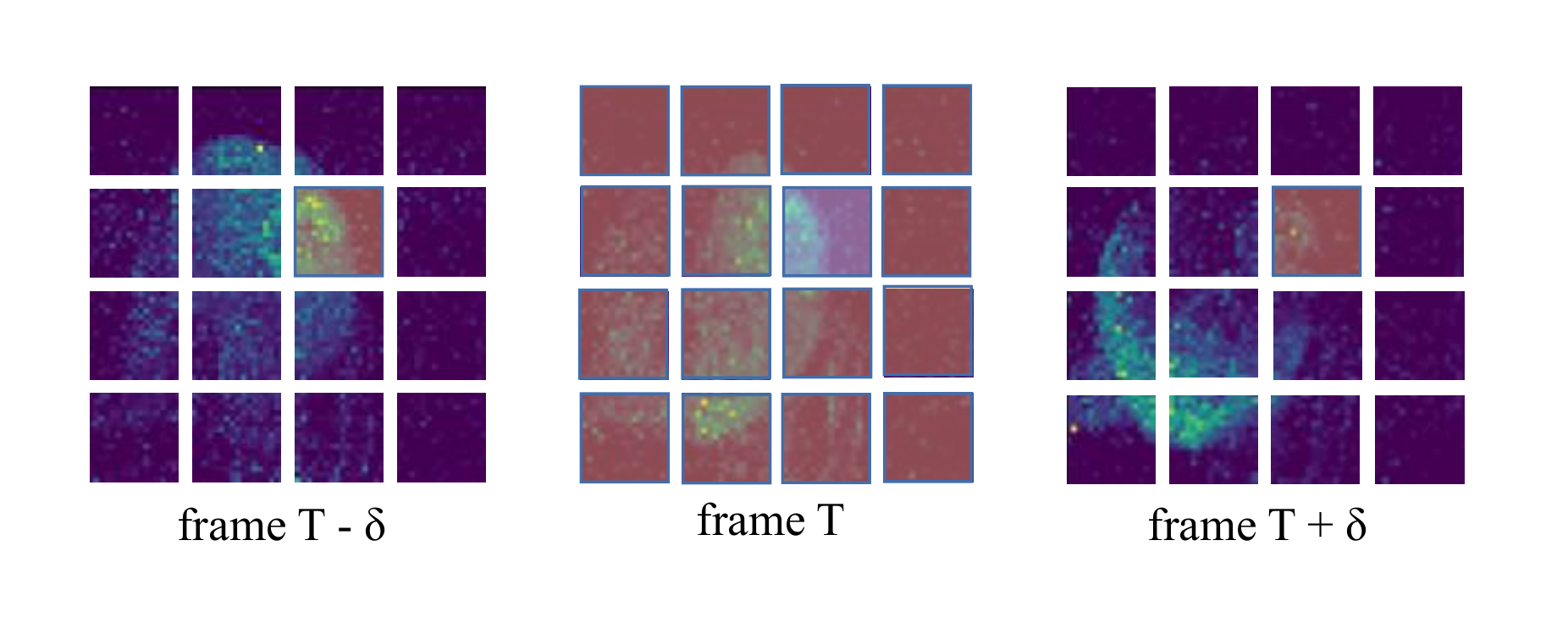}
        \caption{Illustration of Divided Space-Time Attention scheme. We denote in white the 
        query patch and show in orange its self-attention space-time neighborhood. Patches 
        without color are not used for self-attention computation of the white patch.}
        \label{fig4}
    \end{figure}
    
    Intuitively, the divided Space-Time attention can be illustrated by Fig.~\ref{fig4}. 
    The query token in 
    frame T first computes its attention scores with other tokens that are at the same location in 
    other frames as temporal attention. And then it computes its attention scores with other 
    tokens that are in the same frame as spatial attention. Not only can this scheme reduce 
    memory cost and computational complexity, but it also achieves satisfying accuracy.

    \begin{figure}[h]
        \centering
        \includegraphics[width=0.6\columnwidth]{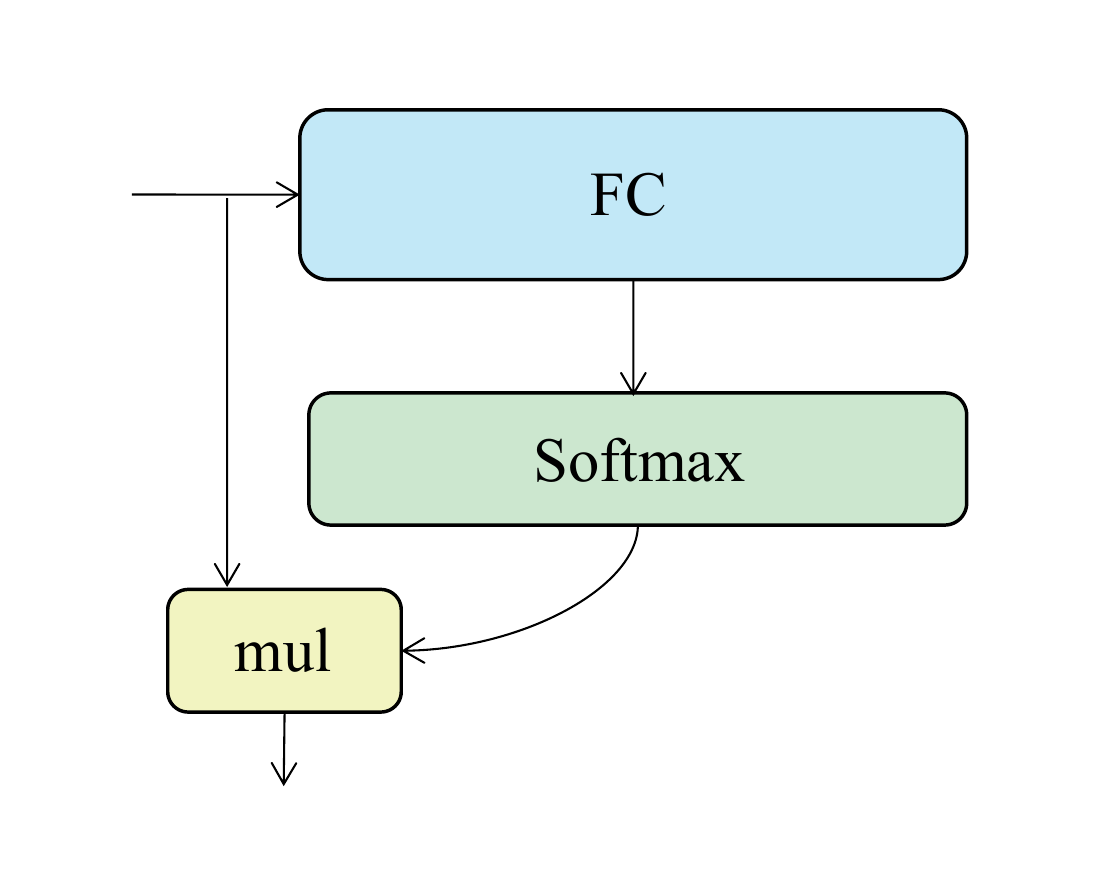}
        \caption{Illustration of Sequence Pooling.}
        \label{fig5}
    \end{figure}
    
    \subsubsection{Sequence Pooling} In order to fully utilize the information in the outputs 
    of the last transformer block, instead of discarding most of them (i.e. class token), we 
    adopt Sequence Pooling to compute the weighted sum of the outputs and send 
    the result to the classifier, as shown in Fig.~\ref{fig5}. Given the normalized output 
    tensor of the last transformer block $z^l \in \mathbb{R}^{T\times N\times D}$, we feed $z^l$ 
    to a fully connected (FC) layer, apply softmax activation, and transpose it:
    \begin{equation}
        W = SM(FC(z^l))^T.
    \end{equation}
    The FC layer is equivalent to a matrix $W_{FC}\in \mathbb{R}^{D\times 1}$ and thus 
    $W\in\mathbb{R}^{T\times 1\times N}$. Then we multiply $z^l$ by $W$:
    \begin{equation}
        \hat{z}=W\times z^l \in \mathbb{R}^{T\times 1\times D},
    \end{equation}
    squeeze the second dimension and get $z\in \mathbb{R}^{T\times D}$. Ultimately, we send the 
    weighted sum $\hat{z}$ to the classifier with a minimal loss of information.
    
    \section{Experiment}
    %82的对比别忘了
    
    We evaluate our models on static dataset (ImageNet) and neuromorphic 
    datasets (DVS-CIFAR10, DVS-Gesture). PLIAF neuron model and PLIF neuron model 
    are adopted for static dataset and neuromorphic datasets, respectively. Extensive ablation 
    experiments reveal the cruciality of CT module and the robustness of our method in terms of 
    depth and time steps. And then we compare our results with state-of-the-art 
    methods and some concurrent works to demonstrate the superiority of our method.
    The details about experimental setup and training strategy 
    are presented in \textbf{Supplementary Material B}.

    \begin{table}[t]
        \centering
        %\resizebox{.95\columnwidth}{!}{
        \begin{tabular}{ccccc}
            \toprule
            \toprule
            Model & \#Layers & \#Heads & Ratio & Dim \\
            \midrule
            Spikeformer-2 & 2 & 2 & 1 & 128 \\
            % \midrule
            Spikeformer-4 & 4 & 2 & 1 & 128 \\
            % \midrule
            Spikeformer-7 & 7 & 4 & 2 & 256 \\
    
            Spikeformer-7L & 7 & 8 & 3 & 512 \\ 
            % \midrule
            % Spikeformer-14& 14& 6 & 3 & 384 \\
            \bottomrule
            \bottomrule
            
        \end{tabular}
        \caption{Details of Spikeformer model variants. Ratio denotes the MLP expansion ratio.}
        \label{table1}
    \end{table}
    
    \subsubsection{Model Variants} We base Spikeformer configurations on those used for 
    CCT \cite{hassani2021escaping}, as summarized in Table~\ref{table1}. We use brief notation to indicate 
    model variants and the CT stem: for instance, Spikeformer-7/5$\times$2$\times$3 means the 
    Spikeformer-7 variant using $2\times3=6$ CT modules with kernel size 5 and 
    every 2 CT modules have 1 downsample module. Note that every downsample CT module 
    will reduce the number of tokens 
    by 4, which will be further discussed in the ablation studies of CT module. 
    Details of the model architecture can be found in \textbf{Supplementary Material C}.
    
    \subsubsection{Datasets} We verify our method on three popular datasets, which consist of 
    both static dataset and neuromorphic datasets. The first is DVS-Gesture \cite{Amir2017ALP}, 
    which contains 1342 records in the training set and 288 examples for testing captured by 
    DVS cameras. The second is DVS-CIFAR10 \cite{li2017cifar10}, which is converted from the 
    famous CIFAR10 dataset to its dynamic form by scanning 
    each sample in front of DVS cameras. The last one is 
    ImageNet \cite{deng2009imagenet}, which is the most popular benchmark dataset used for 
    large-scale image classification.
    
    \begin{table}[t]
        \centering
        \begin{tabular}{ccc}
            \toprule
            \toprule
            CT & SN & Accuracy(\%) \\
    
            \midrule
    
            \XSolidBrush & \XSolidBrush & 82.29 \\
    
            \XSolidBrush & \CheckmarkBold & N/A \\
    
            \CheckmarkBold & \XSolidBrush & 95.83 \\
    
            \CheckmarkBold & \CheckmarkBold & \textbf{98.96} \\
    
            \bottomrule
            \bottomrule
    
        \end{tabular}
        \caption{Ablation studies on the Convolutional Tokenizer (CT) module 
        and spiking neurons (SN) on DVS-Gesture dataset. 
        N/A denotes that the model does not converge.}
        \label{table2}
    \end{table}
    
    \subsection{Ablation Studies}
    We perform extensive ablation experiments to verify the importance of the CT module and draw a 
    empirical rule to design the CT module. And then we conduct experiments with various time 
    steps and layers of transformer block to further demonstrate the robustness of our method. 
    Note that the ablation studies are not aimed at pushing SOTA results, so we adopt 
    realtively simple training recipes.

    \begin{table}[t]
        \centering
        \begin{tabular}{cccc}
            \toprule
            \toprule
            Dataset & Tokens & Accuracy(\%) \\
            
            \midrule
    
            \multirow{6}{*}{DVS-CIFAR10}
    
            & 4096 & N/A \\
    
            & 1024 & 73.10 \\
    
            & 256 & \textbf{80.0} \\
    
            & 64 & 79.6 \\
    
            & 16 & 75.4 \\
    
            & 4 & 73.0 \\
    
            \midrule
    
            \multirow{6}{*}{DVS-Gesture}
    
            & 4096 & N/A \\
    
            & 1024 & 82.64 \\
    
            & 256 & \textbf{90.97} \\
    
            & 64 & 88.89 \\
    
            & 16 & 87.50 \\
    
            & 4 & 85.42 \\
    
            \bottomrule
            \bottomrule
        \end{tabular}
        \caption{Ablation on the CT architecture.
        Note that increasing the number of tokens to 4096 causes a GPU memory overflow.}
        \label{table3}
    \end{table}

    \begin{table*}[t]
        \centering
        %\resizebox{.95\columnwidth}{!}{
        \begin{tabular}{ccccc}
            % \toprule
            % \toprule
            % Work & Model & Time step & Neuron & Parameters(M) & Accuracy(\%) \\
            % \midrule
            
            % (Zheng et al, 2021) & ResNet-17 & 40 & LIF & 1.87 & 96.87 \\
    
            % (Fang et al, 2021) & 5Conv, 2Fc & 20 & PLIF & 1.70 & 97.57 \\
            
            % (Fang et al, 2021) & 7B-Net & 16 & PLIF & 0.13 & 97.92 \\
    
            % \multirow{2}*{(Yao et al, 2021)} & \multirow{2}*{TA-SNN} 
            % & \multirow{2}*{60} & LIF & 1.27 & 95.48 \\ & & & LIAF & 1.27 & 98.61 \\
            
            % This work & Spikeformer-7/5$\times$3 & 20 & PLIF & 9.28 & 98.61 \\
            % \bottomrule
            % \bottomrule
    
            \toprule
            \toprule
            Work & Model & Time Steps & Neuron & Accuracy(\%) \\
            \midrule
            \cite{shen2022eventmix} & ResNet-18 & 10 & PLIF & 96.75 \\
            
            % \cite{zheng2021going} & ResNet-17 & 40 & LIF & 96.87 \\
    
            \cite{fang2021incorporating} & 5Conv, 2Fc & 20 & PLIF & 97.57 \\
            
            \cite{fang2021deep} & 7B-Net & 16 & PLIF & 97.92 \\
    
            \multirow{2}*{\cite{yao2021temporal}} & \multirow{2}*{TA-SNN} 
            & \multirow{2}*{60} & LIF & 95.48 \\ & & & LIAF & 98.61 \\
            
            \textbf{This work} & Spikeformer-7/5$\times$1$\times$3 & \textbf{16} & PLIF & \textbf{98.96} \\
            \bottomrule
            \bottomrule
            
        \end{tabular}
        \caption{Comparison with the SOTA methods on DVS-Gesture dataset.}
        \label{table4}
    \end{table*}
    
    \subsubsection{CT Module} When using the patchify stem, the ANN version performs poorly 
    (82.29\%) on DVS-Gesture. There is a large gap between the training accuracy (100\%) and the 
    testing accuracy (82.29\%). The model is severely overfitted because of the lack of inductive 
    biases. And the training accuracy of the SNN version even does not converge because of the 
    substandard optimizability caused by the patchify stem \cite{xiao2021early}. 
    To deal with it, we propose CT module to help stabilize the training period, 
    alleviate ``data hungry'' and make the model generalize better. In Table~\ref{table2}, we can 
    observe that CT module greatly boosts the accuracy. With CT module, we obtain over 10\% 
    improvement for ANN version and successfully train the SNN version (i.e. Spikeformer), 
    which further enlarges the gap to more than 16\%. These results illustrate the 
    indispensability of the CT module.

    Moreover, the design of the CT module is crucial for the performance of Spikeformer. The more 
    the downsample blocks, the fewer the number of tokens. And the number of tokens 
    determines the granularity of the feature map, 
    which will influence the performance of transformer blocks \cite{dosovitskiy2020image}. 
    From Table~\ref{table3}, we note that Spikeformer performs best with around 200 tokens. Thus, 
    we follow this rule to design our models in the following experiments.

    \subsubsection{Depth And Time Steps} In addition to the CT module, we also conduct 
    depth analysis on transformer blocks. We experiment with three model variants on 
    DVS-Gesture and DVS-CIFAR10 datasets, as shown in Table~\ref{table5}. Table~\ref{table5} shows that the accuracy 
    on both datasets consistently increases with the deepening of transformer blocks. And even 
    with shallow architecture (i.e. Spikeformer-2), our model still performs well.

    \begin{table}[h]
        \centering
        \begin{tabular}{cccc}
            % \toprule
            \toprule

            Dataset & Model & Acc.(\%) \\
    
            \midrule
    
            \multirow{3}{*}{DVS-Gesture}
    
            & Spikeformer-7/5$\times$1$\times$3 & \textbf{98.96} \\
            & Spikeformer-4/5$\times$1$\times$3 & 97.22 \\
            & Spikeformer-2/5$\times$1$\times$3 & 96.88 \\
    
            \midrule
    
            \multirow{3}{*}{DVS-CIFAR10}
    
            & Spikeformer-7/3$\times$2$\times$3 & \textbf{80.0} \\
            & Spikeformer-4/3$\times$2$\times$3 & 78.6 \\
            & Spikeformer-2/3$\times$2$\times$3 & 77.9 \\
    
            \bottomrule
            % \bottomrule
    
        \end{tabular}
        \caption{Ablation on the depth of the transformer blocks.}
        \label{table5}
    \end{table}

    \begin{table}[t]
        \centering
        % \resizebox{0.6\columnwidth}{!}{
        \begin{tabular}{cccc}
            % \toprule
            \toprule
            \multicolumn{2}{c}{DVS-Gesture} & 
            \multicolumn{2}{c}{DVS-CIFAR10} \\
    
            \midrule
    
            T & Acc.(\%) & T & Acc.(\%) \\
    
            \midrule
    
            16 & \textbf{98.96} & 4 & \textbf{78.8} \\
            12 & 97.22 & 3 & 77.9 \\
            8  & 95.14 & 2 & 77.7 \\
            4  & 93.75 & 1 & 77.6 \\
    
            \bottomrule
            % \bottomrule
    
        \end{tabular}
        % }
        \caption{Ablation on time steps.}
        \label{table6}
    \end{table}

    To further demonstrate the robustness of our method, we evaluate our model with 
    different time steps, as shown in Table~\ref{table6}. On DVS-CIFAR10 dataset, our method does not 
    suffer from severe degradation even with extremely low time steps. We attribute this phenomenon 
    to the dataset itself. DVS-CIFAR10 is collected by scanning static images with DVS cameras. 
    Hence, this dataset inherently does not contain much temporal information. Enlarging the 
    time steps is equivalent to making the network vote more times. The performance on 
    DVS-CIFAR10 is more relevant to the ability to capture spatial relationships. 
    Rather, DVS-Gesture dataset contains 
    abundant temporal information because it is a collection of moving gestures performed by 
    different individuals. Thus, its performance is more sensitive to the simulation time steps. 
    The performance on DVS-Gesture is more relevant to the ability to capture temporal 
    relationships. And our method achieves superior results on both datasets with 
    various time steps, which demonstrates the strong spatio-temporal modeling 
    capabilities of our method.

    \begin{table*}[t]
        \centering
        % \resizebox{1.99\columnwidth}{!}{
        \begin{tabular}{cccccc}
            \toprule
            \toprule
            Work & Model & Time Steps & Neuron & Parameters(M) & Accuracy(\%) \\
            \midrule
            
            % \cite{zheng2021going} & ResNet-19 & 10 & LIF & 14.66 & 67.8 \\
    
            \cite{fang2021incorporating} & 4Conv, 2Fc & 20 & PLIF & 17.4 & 74.8 \\
            
            \cite{fang2021deep} & Wide-7B-Net & 4, 8, 16 & PLIF & 1.19 & 64.8, 70.2, 74.4 \\
    
            \multirow{2}*{\cite{yao2021temporal}} & \multirow{2}*{TA-SNN} 
            & \multirow{2}*{10} & LIF & 2.10 & 71.1 \\ & & & LIAF & 2.10 & 72.0 \\
    
            \cite{li2021differentiable} & ResNet-18 & 10 & LIF & 11.69 & 75.4 \\
            
            \cite{meng2022training} & VGG-11 & 20 & LIF & 132.86 & 77.3 \\
    
            \cite{shen2022eventmix} & ResNet-18 & 10 & PLIF & 11.69 & 81.45 \\ 
    
            \cite{li2022neuromorphic} & VGG-11 & 10 & LIF & 132.86 & 81.7 \\
    
            \textbf{This work} & Spikeformer-7/3$\times$2$\times$3 & \textbf{4} & PLIF & \textbf{9.28} & 
            \textbf{81.4} \\
            \bottomrule
            \bottomrule
            
        \end{tabular}
        % }
        \caption{Comparison with the SOTA methods on DVS-CIFAR10 dataset.}
        \label{table7}
    \end{table*}

    \begin{table*}[t]
        \centering
        % \resizebox{2.0\columnwidth}{!}{
        \begin{tabular}{cccccc}
            \toprule
            \toprule
            % Method & Work & Model & Time step & Neuron & Parameters(M) & Accuracy(\%) \\
            Method & Work & Model & T & Par.(M) & Accuracy(\%) \\
    
            \midrule
            \multirow{5}{*}{\textbf{ANN}} &
    
            \multirow{3}*{(He et al, 2015)} 
            % & ResNet-18 & - & 11.69 & 72.12 \\ & 
            % & ResNet-34 C & - & 21.80 & 75.81 \\ & 
            & ResNet-50 & - & 25.56 & 77.15 \\ & 
            & ResNet-101 & - & 44.55 & 78.25 \\ & 
            & ResNet-152 & - & 60.20 & 78.57 \\ &
    
            \multirow{2}*{\cite{beyer2022better}} 
            & ViT-S/16* & - & \multirow{2}*{22.04} & 67.1 \\ &
            & ViT-S/16$\dag$ & - &  & 76.1 \\ 
            % &
            % & ViT-S/16 & - & - &  & 80.0 \\
    
            \midrule
            \multirow{4}*{\textbf{ANN-to-SNN}} & 
    
            \cite{sengupta2019going} & VGG-16 & 2500 & 138.36 & 69.96 \\ &
    
            % (Han et al. 2020) & ResNet-34 & 1024 & IF & 21.80 & 66.61 \\ &
    
            \cite{hu2018spiking} & ResNet-50 & 350 & 25.56 & 72.75 \\ &
    
            \cite{han2020rmp} & VGG-16 & 4096 & 138.36 & 73.09 \\
    
            \midrule
    
            \multirow{5}*{\textbf{Directly Training}} &
            
            \cite{zheng2021going} & ResNet-34(large) & 6 & 86.13 & 67.05 \\ &
    
            \cite{fang2021deep} & SEW ResNet-152 & 4 & 60.19 & 69.26 \\ &
    
            \cite{li2021differentiable} & VGG-16 & 5 & 138.36 & 71.24 \\ &
            
            \cite{meng2022training} & PreAct-ResNet-18 & 50 & 11.69 & 67.74 \\ &
    
            % \cite{hu2021advancing} & ResNet-104 & 5 &  & 74.21 \\ &
    
            \textbf{This work} & Spikeformer-7L/3$\times$2$\times$4 & \textbf{4} & 
            \textbf{38.75} & \textbf{78.31} \\

            \bottomrule
            \bottomrule
            
        \end{tabular}
        % }
        \caption{Comparison with the SOTA methods on ImageNet dataset.*The original version without 
        additional data and strong data augmentation. $\dag$The original version without additional 
        data but adopts strong data augmentation.}
        \label{table8}
    \end{table*}

    \subsection{Comparison With SOTA Methods}
    % 更好的利用有限的参数来表征
    Following the ``200 tokens'' rule, we adopt three different architectures. 
    Since the input resolution of DVS-Gesture and DVS-CIFAR10 is $128\times 128$, we downsample 
    the input frames for 3 times, which results in 256 tokens. And for ImageNet with an input size 
    of $224\times 224$, we downsample the input for 4 times, which results in 196 tokens. 
    Because DVS-Gesture dataset contains rich temporal information, it needs more time steps 
    to gain better performance and takes up more memory. To balance the memory cost, we 
    adopt Spikeformer-7/5$\times$1$\times$3 for DVS-Gesture and Spikeformer-7/3$\times$2$\times$3 
    for DVS-CIFAR10. But the original Spikeformer-7 variant is too simple for ImageNet dataset, 
    we adopt Spikeformer-7L/3$\times$2$\times$4 for ImageNet and compare these methods 
    with SOTA methods and concurrent works.

    \subsubsection{DVS-Gesture} 
    In Table~\ref{table4}, we compare our method with SOTA methods on DVS-Gesture. 
    \cite{fang2021deep} achieve 97.92\% top-1 accuracy with 16 time steps, while we 
    obtain 98.96\% with the same time steps. 
    Compared to \cite{yao2021temporal}, we achieve better result with much lower latency 
    (i.e. less than one-third of their simulation time steps) and spike-based neurons (i.e. lower energy consumption). 
    Since the performance on DVS-Gesture tends to saturate, the capacity of our 
    method is not fully demonstrated. Thus we further conduct experiments on the 
    more challenging DVS-CIFAR10 dataset.

    \subsubsection{DVS-CIFAR10} Our method achieves remarkable result on DVS-CIFAR10, as shown in 
    Table~\ref{table7}. With the same time steps as \cite{fang2021deep}, our method outperforms their results by 
    a large margin (i.e. 16.6\%) and even achieves better performance than \cite{meng2022training}, who 
    use five times as many time steps as ours and ten times more parameters. Our method gains 
    comparable results to \cite{li2022neuromorphic} and \cite{shen2022eventmix} with fewer time steps and parameters. 
    These comparisons 
    strongly support that our Spikeformer is equipped with distinguished spatio-temporal 
    representational ability that can fully utilize limited spatio-temporal information to 
    extract key information.

    \subsubsection{ImageNet} To explore the potential of Spikeformer to be a large-scale SNN 
    architecture, we conduct experiments on ImageNet dataset with PLIAF model, as shown in Table~\ref{table8}. Our 
    Spikeformer outperforms other SNNs trained by SG method or conversion method by a large margin 
    with low latency and medium parameters. Moreover, we even outperform ViT-S/16 that is not 
    pretrained with additional data, which further demonstrates that our CT module do help 
    alleviate ``data hungry''. To our best knowledge, this is the first time directly trained 
    SNNs achieve comparable results to ANNs on ImageNet with fewer parameters. 
    Given limited time steps and parameters, 
    our model manages to focus on the most informative components of 
    the input and possesses the capability to fully utilize parameters to learn robust 
    representation. 
    These results indicate that Spikeformer is a promising architecture for training large-scale SNNs.
    
    \section{Conclusion}
    In this paper, we introduce spatio-temporal attention (STA) into spiking neural network (SNN) 
    and directly train a Transformer-based SNN (i.e. Spikeformer). 
    Replacing the original patchify stem with CT module, we manage to alleviate ``data hungry'', 
    stabilize the training period and obtain a high-performance 
    low-latency SNN.
    Based on extensive ablation studies, we demonstrate the cruciality of CT module 
    and the robustness of Spikeformer in terms of depth and time steps. 
    Furthermore, 
    we also propose an empirical rule to design CT module (i.e. the ``200 tokens'' rule). This 
    module can be further 
    developed into a delicately designed multi-stages network (e.g. ResNet) in future work. 
    But in this paper, we adopt simple design as not to overwhelm the importance of 
    transformer blocks, which may be a suboptimal choice. 
    Ultimately, 
    our Spikeformer gains competitive or state-of-the-art results 
    on both neuromorphic datasets and 
    large-scale static dataset with low latency, 
    which demonstrates its superior capabilities to capture spatio-temporal information and 
    the potential to be an alternative architecture to CNN for training large-scale SNNs. 
    We believe this architecture will bring new insight into training large-scale SNNs and 
    promote the development of SNNs.    

\bibliography{aaai22}
\section{Supplementary Material}

\setcounter{secnumdepth}{1} %May be changed to 1 or 2 if section numbers are desired.

\appendix

\section{Methods and Theoretical Analysis}

\subsection{Methods}

For all experiments, we adopt the same surrogate gradient function to approximate the 
gradients of the non-differentiable activation function on the backpropagation process. 
The function can be formulated as follows: 
\begin{equation}
    \sigma(x) = \frac{1}{\pi}\arctan{(\frac{\pi}{2}\alpha x)} + \frac{1}{2},
\end{equation}
\begin{equation}
    \sigma'(x) = \frac{\alpha}{2(1+(\frac{\pi}{2}\alpha x)^2)},
\end{equation}

\begin{figure}[h]
    \centering
    \includegraphics[width=0.9\columnwidth]{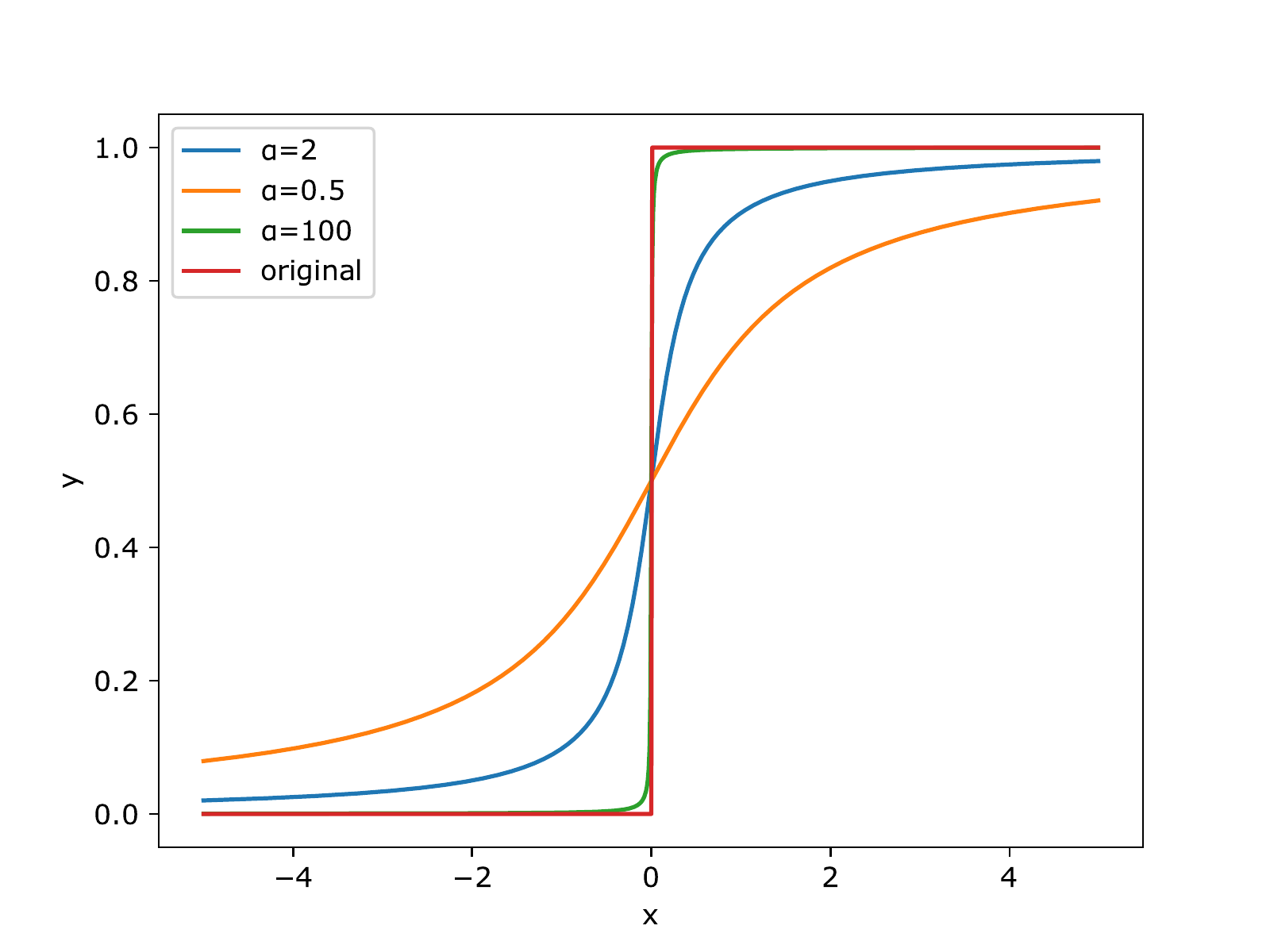}
    \caption{Visualization of the surrogate function and the original activation function.}
    \label{fig6}
\end{figure}

\begin{figure*}[t]
    \centering
    \subfigure[Original block in ResNet]{
    % \begin{minipage}{7cm}
    \centering
    \includegraphics[width=0.67\columnwidth]{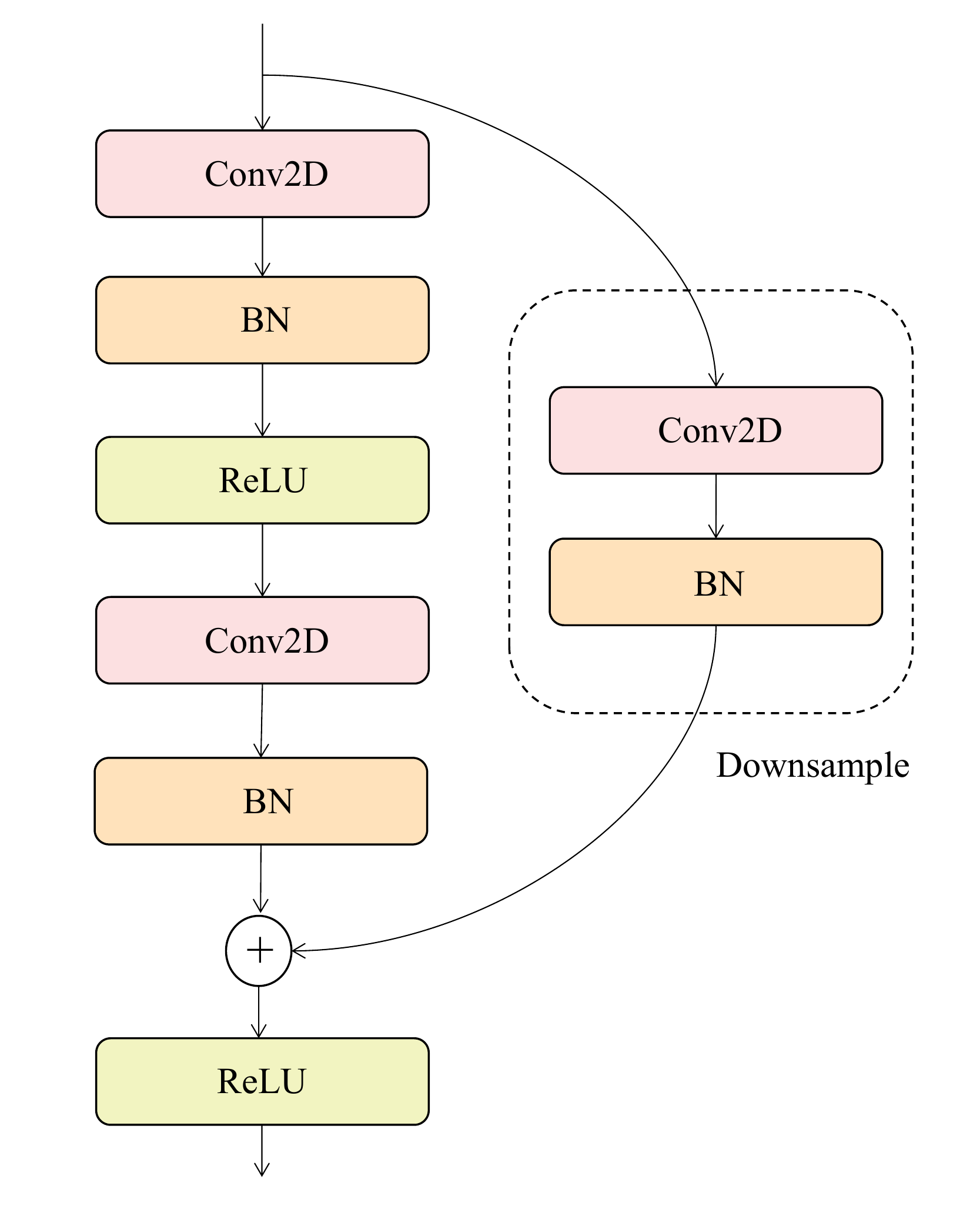}
    % \end{minipage}
    }
    \hspace{1in}
    \subfigure[block in CT module]{
    % \begin{minipage}{7cm}
    \centering
    \includegraphics[width=0.6\columnwidth]{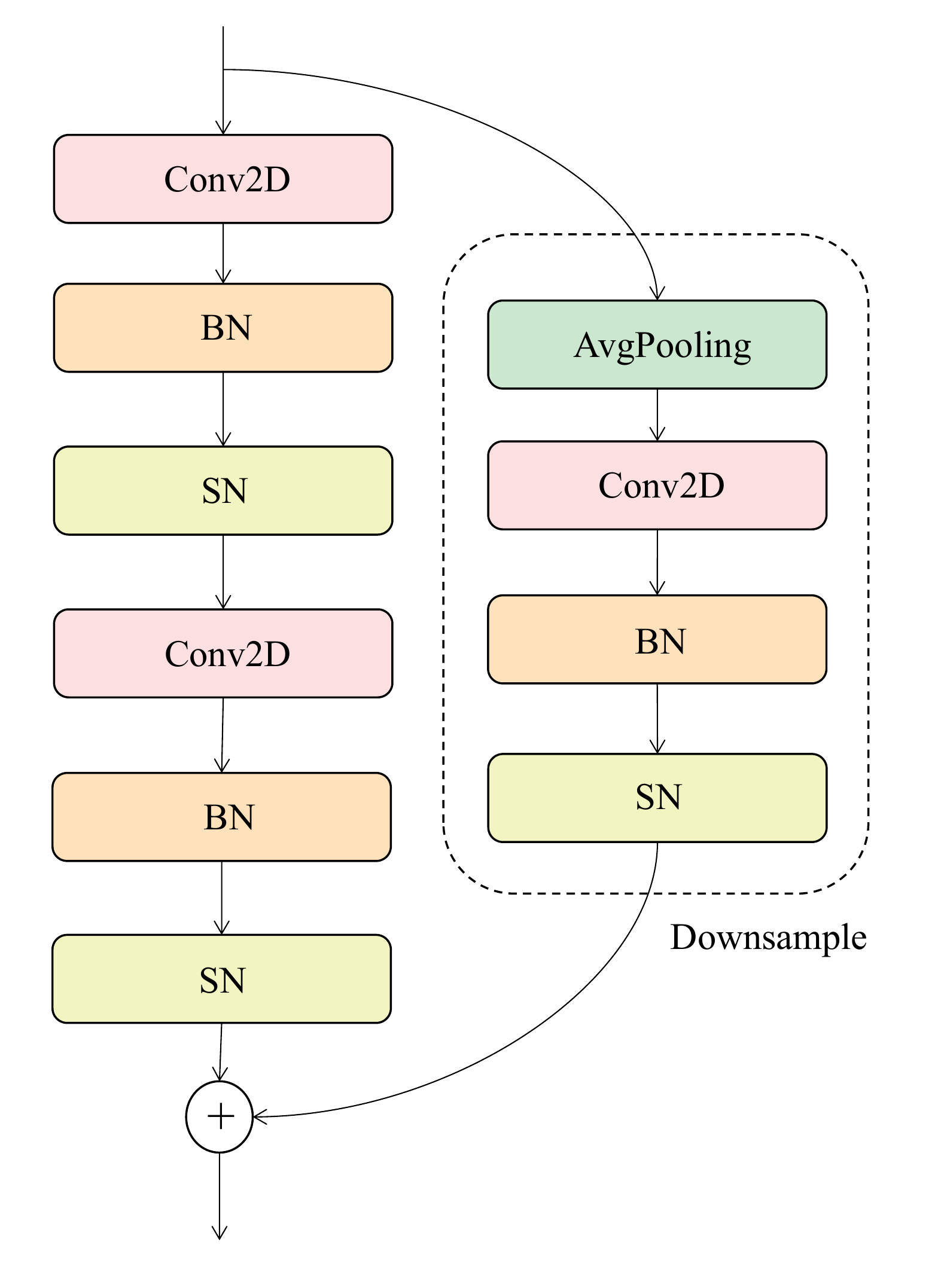}
    % \end{minipage}
    }
    \caption{Comparison of the original block in ResNet and the 
    block in CT module.}
    \label{fig7}
\end{figure*}

where $\alpha$ is the slope parameter controlling how steep the function is. This parameter 
can be further set as a learnable parameter, but this is out of the scope of this paper. 
We set $\alpha=2$, $V_{reset}=0$ and $V_{th}=1$ for all neurons. We detach $S[t]$ in 
% Eq.~\eqref{eq2} 
Eq.~(2) 
in the backward computational graph to improve performance \cite{zenke2021remarkable}. 
The original activation function and surrogate gradient function are visualized in Fig.~\ref{fig6}. 
Note that with the increase of the value of $\alpha$, the surrogate function becomes 
steeper. The activation function works fine during the forward propagation, but the surrogate 
function takes its place in the computational graph. And thus, we manage to backpropagate 
the gradients of the non-differentiable activation function.

We adopt mixed precision training \cite{micikevicius2018mixed} to accelerate training 
and save memory, but it may result in slightly lower accuracy. For ImageNet, we train our 
model on multi-GPU and adopt sync-BN technique \cite{zhang2018context} to reduce the 
influence of small batch size. Additionally, we adopt Droppath \cite{huang2016deep} 
regularization with hyperparameter set to 0.1 for all experiments.

\subsection{Theoretical Analysis}
The comparison of the original block in ResNet \cite{he2016deep} and the 
block in CT module is shown in Fig.~\ref{fig7}. As discussed by \cite{fang2021deep}, the 
original design does not fit into SNN. Take the basic block without downsampling as an example: 
\begin{equation}
    X^{l + 1} = ReLU(F^l(X^l) + X^l),
    \label{eq21}
\end{equation}
\begin{equation}
    X^{l + 1} = SN(F^l(X^l) + X^l),
    \label{eq22}
\end{equation}
where $X^l$ and $F^l(\cdot)$ denote the input and the residual mapping in the $l^{th}$ block. 
Eq.~\eqref{eq21} and Eq.~\eqref{eq22} are the ANN version and SNN version respectively. 
Assuming that the blocks are zero-initialized (i.e. all weights equal to zero), 
we have:
\begin{equation}
    X^{l + 1} = ReLU(X^l),
    \label{eq23}
\end{equation}
\begin{equation}
    X^{l + 1} = SN(X^l).
    \label{eq24}
\end{equation}
In Eq.~\eqref{eq23}, $X^l$ has been processed by the $ReLU$ function of the $l-1^{th}$ block, and 
thus $X^{l+1} = X^l$, which is identity mapping. As for Eq.~\eqref{eq24}, due to the leaky 
property of most spiking neurons, it is hard to implement identity mapping unless we 
change the neural dynamics equation 
% Eq.~\eqref{eq1} 
Eq.~(1) 
into:
\begin{equation}
    H[t] = V[t - 1] + X[t],
    \label{eq25}
\end{equation}
and assign a small enough value to $V_{th}$, which enables the neurons to output a spike as 
soon as they receive a spike. But this compromise limits the applications of SNNs, so the 
original design of ResNet is not suitable for SNNs. To deal with it, \cite{fang2021deep} 
propose SEW block to achieve identity mapping. We adopt $ADD$ as element-wise operation 
and obtain:
\begin{equation}
    X^{l + 1} = SN(F^l(X^l)) + X^l.
\end{equation}
Spikes are propagating through these blocks. With zero-initialized, it can easily implement 
identity mapping. Because the $ReLU$ function will not shrink input, this architecture does 
not fit into its ANN counterpart. But the output of spiking neurons is 0/1, so every time 
there is a downsample block, the input will be no greater than 2, which effectively tackles 
this issue. Besides, we modify the shortcut of downsample block as stride-2 
convolutional layer with kernel size 
1 may lose information \cite{he2019bag}. So we use a $2\times 2$ average pooling layer with 
a stride of 2 to downsample the feature map and a stride-1 convolutional layer with 
kernel size 1 to expand the dimension. The average pooling layer integrates inputs and 
transfers them to the following layers. It will not hinder the transfer of information and 
have a minimal impact on identity mapping. Further discussions and experiments can be found 
in \cite{fang2021deep}.

\begin{table*}[t]
    \centering
    % \resizebox{2.0\columnwidth}{!}{
        % >{$}  <{$}
    \begin{tabular}{c|c|c|c|c}
        \hline
        & Spikeformer-4/5$\times$1$\times$3 
        & Spikeformer-7/5$\times$1$\times$3 & Spikeformer-7/3$\times$2$\times$3 & 
        Spikeformer-7L/3$\times$2$\times$3\\

        \hline

        block1 & 
        $\begin{pmatrix} 5\times 5, 32 \\ 5\times 5, 32 \end{pmatrix}$ 
        $\times$ 1 & 
        $\begin{pmatrix} 5\times 5, 64 \\ 5\times 5, 64 \end{pmatrix}$ 
        $\times$ 1 & 
        $\begin{pmatrix} 3\times 3, 64 \\ 3\times 3, 64 \end{pmatrix}$ 
        $\times$ 2 & 
        $\begin{pmatrix} 3\times 3, 64 \\ 3\times 3, 64 \end{pmatrix}$ 
        $\times$ 2 \\

        \hline

        block2 & 
        $\begin{pmatrix} 5\times 5, 64 \\ 5\times 5, 64 \end{pmatrix}$ 
        $\times$ 1 & 
        $\begin{pmatrix} 5\times 5, 128 \\ 5\times 5, 128 \end{pmatrix}$ 
        $\times$ 1 & 
        $\begin{pmatrix} 3\times 3, 128 \\ 3\times 3, 128 \end{pmatrix}$ 
        $\times$ 2 & 
        $\begin{pmatrix} 3\times 3, 128 \\ 3\times 3, 128 \end{pmatrix}$ 
        $\times$ 2 \\
        
        \hline
        
        block3 & 
        $\begin{pmatrix} 5\times 5, 128 \\ 5\times 5, 128 \end{pmatrix}$ 
        $\times$ 1 & 
        $\begin{pmatrix} 5\times 5, 256 \\ 5\times 5, 256 \end{pmatrix}$ 
        $\times$ 1 & 
        $\begin{pmatrix} 3\times 3, 256 \\ 3\times 3, 256 \end{pmatrix}$ 
        $\times$ 2 & 
        $\begin{pmatrix} 3\times 3, 256 \\ 3\times 3, 256 \end{pmatrix}$ 
        $\times$ 2 \\

        \hline

        block4 & 
        \diagbox[dir=SW, 
        innerwidth=\widthof{ Spikeformer-7/5$\times$1$\times$3 }, 
        height=2\line]{}{}
        & \diagbox[dir=SW, 
        innerwidth=\widthof{ Spikeformer-7/5$\times$1$\times$3 }, 
        height=2\line]{}{}
        & \diagbox[dir=SW, 
        innerwidth=\widthof{ Spikeformer-7/5$\times$1$\times$3 }, 
        height=2\line]{}{}
        & 
        $\begin{pmatrix} 3\times 3, 512 \\ 3\times 3, 512 \end{pmatrix}$ 
        $\times$ 2 \\

        \hline
        
    \end{tabular}
    % }
    \caption{Architectures of the CT module.}
    \label{table9}
\end{table*}

\section{Details of Experiments}
\subsection{DVS-Gesture}
The input resolution of DVS-Gesture is $128\times 128$. We use the same AER data pre-processing 
method as \cite{fang2021incorporating} and do not adopt any data augmentation for this 
dataset. We train our model using Adam \cite{kingma2015adam} with $\beta_1=0.9$, 
$\beta_2=0.999$, a batch size of 16, and apply a weight decay of 0.00015. We warm up 
for 20 epochs and train the model for a total of 150 epochs with a learning rate of 0.001. 

\subsection{DVS-CIFAR10}
The input resolution and the AER data pre-processing method of DVS-CIFAR10 are the same 
as DVS-Gesture. Because DVS-CIFAR10 is more challenging than DVS-Gesture, we adopt additional 
data augmentation. First, we horizontally flip the input frames with a 
possibility of 0.5. Then we sample two values $a$ and $b$ from an integer uniform distribution 
$U(-5, 5)$, horizontally move the input frames for $a$ pixels, and vertically move the 
input frames for $b$ pixels. When $a$/$b$ is positive, the frames are moved upwards/rightwards, 
and vice versa. Finally, we randomly select two values $l$ and $h$ from an integer uniform 
distribution $U(1, 16)$ as length and height, mask off an area of 
the frames with the selected $l$, and $w$, and adopt label-smoothing regularization 
\cite{szegedy2016rethinking} with the hyperparameter set to 0.14. 
We train our model using Adam \cite{kingma2015adam} 
with $\beta_1=0.9$, $\beta_2=0.999$, a batch size of 32, and apply a weight decay of 0.0001. 
We warm up for 25 epochs and train the model for a total of 600 epochs with an initial 
learning rate of 0.001. Every 192 epochs, we decay the learning rate from $lr$ to $0.1lr$. 

\subsection{ImageNet}
We first randomly crop the images to $224\times 224$ and horizontally flip the images with 
a possibility of 0.5. Auto-Augment \cite{Cubuk_2019_CVPR} and label-smoothing 
regularization with the hyperparameter set to 0.1 are adopted for further augmentation. 
We train our model using SGD with 
a momentum of 0.9, a total batch size of 160 on 8 GPUs, and apply a weight decay of 0.0001. 
We warm up for 5 epochs and train the model for a total of 130 epochs with an initial 
learning rate of 0.01 and a cosine learning rate decay scheduler \cite{loshchilov2017sgdr}. 
Besides, we manually decay the learning rate by 0.1 when the training accuracy tends to 
plateau (at epoch \{94, 115, 118\} respectively).

\section{Model Architecture}
The detailed architectures of CT module are shown in Table~\ref{table9}. The first 
convolutional layer of each block will downsample the input with a stride of 2 and 
each block will double the dimension until it reaches the dimension of transformer block.

\section{Reproducibility}
Our implementations are based on PyTorch \cite{paszke2019pytorch} and 
SpikingJelly \cite{SpikingJelly}. Our code will be publicly available, and for 
reproducibility, we use the identical seed in all experiments and will detail our 
configurations in our code repository.

\end{document}